\documentclass{article}

\usepackage{PRIMEarxiv}

\usepackage[utf8]{inputenc}
\usepackage{newunicodechar}
\newunicodechar{−}{\ensuremath{-}}
\usepackage{natbib} 

\usepackage[utf8]{inputenc} 
\usepackage[T1]{fontenc}    
\usepackage{hyperref}       
\usepackage{url}            
\usepackage{booktabs}       
\usepackage{amsfonts}       
\usepackage{nicefrac}       
\usepackage{microtype}      
\usepackage{xcolor}         
\usepackage{graphicx}
\usepackage{subcaption}
\usepackage{enumitem}
\usepackage{amsmath}
 \usepackage{wrapfig}
 \usepackage{tcolorbox}    
\tcbuselibrary{breakable}
\usepackage{subcaption} 

\usepackage[utf8]{inputenc} 
\usepackage[T1]{fontenc}    
\usepackage{hyperref}       
\usepackage{url}            
\usepackage{booktabs}       
\usepackage{amsfonts}       
\usepackage{nicefrac}       
\usepackage{microtype}      
\usepackage{xcolor}         

\title{Local Coherence or Global Validity? Investigating RLVR Traces in Math Domains}

\author{%
Soumya Rani Samineni\\
	SCAI, Arizona State University\\
	\texttt{ssamine4@asu.edu} \\
	\And
    Durgesh Kalwar\\
	SCAI, Arizona State University\\
	\texttt{dkalwar@asu.edu} \\
	\And
	Vardaan Gangal \\
	SCAI, Arizona State University\\
	\texttt{vgangal3@asu.edu } \\
    \And
  Siddhant Bhambri\\
	SCAI, Arizona State University\\
	\texttt{siddhantbhambri@asu.edu } \\
	\And
	Subbarao Kambhampati \\
	SCAI, Arizona State University \\
	\texttt{rao@asu.edu} \\
}

\begin{document}

\maketitle

\begin{abstract}

Reinforcement Learning with Verifiable Rewards (RLVR)-based post-training of Large Language Models (LLMs) has been shown to improve accuracy on reasoning tasks and continues to attract significant attention. Existing RLVR methods, however, typically treat all tokens uniformly without accounting for token-level advantages. These methods primarily evaluate performance based on final answer correctness or Pass@K accuracy, and yet make claims about RL post-training leading to improved reasoning traces. This motivates our investigation into the effect of RL post-training on intermediate tokens which are not directly incentivized. To study this, we design an experimental setup using the GRPO algorithm with Qwen-2.5-0.5B model on the GSM8K dataset. We introduce trace coherence, a First-Order Logic (FOL)-based measure to capture the consistency of reasoning steps by identifying errors in the traces. We distinguish between trace validity and trace coherence, noting that the former implies logical soundness while the latter measures local coherence via lack of errors. Our results show that RL post-training overall improves trace coherence with the most significant gains on problems where the base model fails but the RL model succeeds. Surprisingly, RL enhances local coherence without necessarily producing valid or correct solutions. This highlights a crucial distinction: improved local coherence in reasoning steps does not guarantee final answer correctness. We argue that claims of improved reasoning via RL must be examined with care, as these may be based on improved trace coherence, which may not translate into fully valid mathematical proofs.

\end{abstract}

\section{Introduction}

Following the release of Deepseek R1\cite{guo2025deepseek}, post-training Large Language Models (LLMs) using Reinforcement Learning with Verifiable Rewards (RLVR) has gained widespread attention. Since then several works have expanded on reinforcement learning based post-training by altering the loss function, modifying advantage estimation, and utilizing base model resets \citep{liu2025understanding, zhang2025grpo,yu2025dapo, liu2025prorl}. However, recent analysis by \cite{samineni2025rl} highlights structural limitations of current RLVR approaches, particularly due to the uniform distribution of advantages across all tokens. Further, \cite{yue2025does} argued that the accuracy of RLVR models cannot surpass that of the base model demonstrating empirically that Pass@K accuracy drops relative to the base model as K increases. Also, \cite{cui2025entropy} shows that the performance gains can be predicted using entropy of the base model. 

However, these works primarily focus on the limitations of RLVR in terms of final answer accuracy and do not examine its effect on intermediate tokens, or {\em reasoning traces}. Since RLVR verifies only the final answer and distributes rewards uniformly across all tokens, its impact on the reasoning process at the token level lacks investigation. While there have been frequent claims that RLVR improves reasoning, the effect of integrating verifier signals during training on the structure and quality of reasoning traces has not been formally studied.

{\em Since formal verification of intermediate reasoning steps is not tractable at scale, we cannot directly evaluate trace validity.} Instead, we introduce a proxy metric called \textbf{trace coherence}, which reflects the consistency of reasoning steps. It is important to note here that while trace validity implies coherence, the opposite may not be true., Particularly, we measure trace coherence by analyzing the presence of errors (or lack thereof) in the reasoning steps, where errors are defined using a First-Order Logic (FOL) framework (\S \ref{sec:methodology}). To experimentally analyze this problem, we design an experimental setup (\S \ref{sec:expts}) to study the effect of RLVR on traces using a mathematical reasoning benchmark, particularly GSM8K \cite{cobbe2021training}.

Our results (\S \ref{sec:results}) show that RLVR surprisingly improves trace coherence across Pass@K evaluations, especially on problems where the base model fails but the RL-trained model produces a correct final answer. These findings highlight a key distinction that RL post-training can improve local coherence in reasoning traces, captured through error patterns while not guaranteeing the correctness or full trace validity. This distinction is important for interpreting the effects of RLVR on reasoning quality beyond final answer accuracy.


\section{Related Work}
\label{sec:related_work}
Reasoning traces have been studied since the release of DeepSeek R1\cite{guo2025deepseek} to better interpret LLMs and their improved task performance. Thoughtology \cite{marjanovic2025deepseek} provides a systematic analysis of the length, structure, and content of traces generated by DeepSeek R1, focusing on interpretability and safety. However, \cite{bhambri2025cognitively} show that optimizing traces for user interpretability can reduce model performance, revealing a trade-off between interpretability and LLM's task performance. In addition, \cite{Stechly2025BeyondST,bhambri2025interpretable} demonstrate a lack of correlation between trace validity and final answer correctness during Supervised Fine-Tuning (SFT) on maze problems where it is feasible to check trace validity because of formal verifiers. However, trace validity or trace coherence in the context of RLVR post-training, especially for mathematical reasoning tasks, remains underexplored due to challenges in trace analysis. In this work, we systematically address these challenges by defining FOL-based error tags and LLM-as-a-judge \cite{gu2024survey} to evaluate trace coherence.

In early evaluations of LLM performance using Chain-of-Thought (CoT) traces, various taxonomies of errors specific to mathematical reasoning were introduced~\cite{lewkowycz2022solving, wei2022chain, li2024evaluating}, which are summarized in Table~\ref{table1}. These categories have been widely used to identify reasoning mistakes and systematically evaluate LLM performance.


\begin{table}[h]
\centering
\caption{Comparison of error categorizations for intermediate trace analysis across prior works.}
\label{table1}
\resizebox{\columnwidth}{!}{%
\begin{tabular}{@{}lll@{}}
\toprule
\textbf{Minerva CoT (\cite{lewkowycz2022solving})} &
  \textbf{Chain-of-Thought (\cite{wei2022chain})} &
  \textbf{Examiner (\cite{li2024evaluating})} \\ \midrule
\begin{tabular}[c]{@{}l@{}}Incorrect Data, Format Error, \\ Incorrect Calculation, Misunderstands \\ Question, Incorrect Reasoning, \\ Solution Too Short, Hallucinated, \\ Repeats Question, Other Mistakes\end{tabular} &
  \begin{tabular}[c]{@{}l@{}}Calculator Error, Symbol Mapping \\ Error, One Step Missing, Semantic \\ Understanding Error, Incoherent \\ Chain-of-Thought\end{tabular} &
  \begin{tabular}[c]{@{}l@{}}Calculation Error, Counting Error, \\ Context Value Error, Hallucination, \\ Unit Conversion Error, Operator \\ Error, Formula Confusion Error, \\ Missing Step, Contradictory Step\end{tabular} \\ \bottomrule
\end{tabular}%
}
\end{table}


While these error categories have been effective for evaluation and improve the reasoning performance of LLMs through error identification, they often overlap and are neither mutually exclusive nor exhaustive, limiting their use for formally defining trace coherence. To address this, we propose a new set of error categories, motivated by First-Order Logic and designed to be mutually exclusive to study trace coherence (see Table~\ref{tab:error_categories}).

\section{Methodology}
\label{sec:methodology}
\subsection{Error Categories for Mathematical Reasoning}

\begin{table}[htbp]
\centering
\caption{Error categories and their subtypes.}
\label{tab:error_categories}
\resizebox{\columnwidth}{!}{%
\begin{tabular}{@{}l|l@{}}
\toprule
\textbf{Error Category} & \textbf{Description / Subtypes}       \\ \midrule
False Premise &
  \begin{tabular}[c]{@{}l@{}}Conceptual Misunderstanding; Semantic Error; False Assumption; \\ Units Misinterpretation; Incorrect Derivation Step from problem description;\end{tabular} \\
False Rule &
  \begin{tabular}[c]{@{}l@{}}Type / Operand Mismatch; Inference Violation; Operation \\ Misapplication; Quantifier Misuse; Missing Necessary Steps\end{tabular} \\
Calculator Error        & Simple numeric or arithmetic mistakes \\
Format Error &
  Final answer not formatted as required, e.g., missing  \textit{\textbackslash boxed} \\ \bottomrule
\end{tabular}%
}
\end{table}

Mathematical reasoning problems, particularly popular benchmarks like GSM8K, typically require between two and eight steps to solve. These problems primarily involve performing a sequence of elementary calculations using basic arithmetic operations (\(+\), \(-\), \(\times\), \(\div\)) to arrive at the final answer. Each problem can be formalized as comprising: assumptions or facts, an expression or formula to compute the answer, and intermediate steps to produce quantities essential for deriving the final answer. This structure naturally aligns with the construction of FOL representations. Based on the error categorization, we define a set of error types with clear distinctions between them shown in \ref{tab:error_categories}.

\newpage

    
    

\begin{wrapfigure}{r}{0.48\textwidth}  
    \centering
    \includegraphics[width=0.45\textwidth]{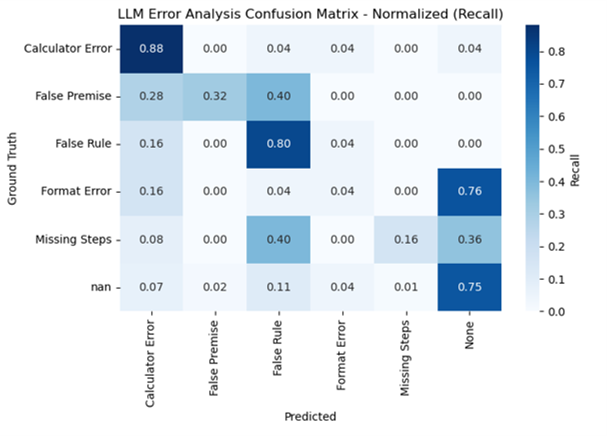}
    \caption{Evaluation accuracy}
    \label{fig:eval}
\end{wrapfigure}

\subsection{Trace Coherence and Pass@K Evaluation Using LLMs}
To assess trace coherence, we first apply our FOL-based error categories to model responses generated during RLVR post-training. We leverage the translative capabilities of LLMs, particularly prompt GPT-4o to convert each response into a FOL representation and classify errors according to our taxonomy(see Appendix~\ref{appendix:error} for details).

Following previous work~\cite{yue2025does}, which evaluated the final accuracy of RL post-trained models across different values of Pass@K, we extend the analysis to measure trace coherence at different Pass@K in addition to accuracy. For each value of k, we define the metrics as follows:

\textbf{Pass@K Accuracy:} A trace is considered correct if \emph{any} of the k responses is correct, and incorrect if \emph{all} k responses are incorrect.
    
\textbf{Pass@K Trace Coherence:} We check for coherence only when the answers are correct, consider c are correct out of k responses. A trace is considered coherent if \emph{at least one} of the c responses is error free, and incoherent otherwise.

    


\section{Experimental Setup}
\label{sec:expts}
\textbf{Datasets:} We perform our analysis on the \textit{GSM8K} dataset, a widely used benchmark of grade-school math problems. It contains 8.5K problems, each comprising a question and its corresponding answer, divided into 7.5K training problems and 1K test problems.

\textbf{Base Model:} Our experiments use Qwen-2.5-0.5B from the Qwen-2.5 family as the base model. The model is fine-tuned using the VERL pipeline \cite{sheng2025hybridflow}. See training hyperparameters in appendix~\ref{appendix:hyper_params}.


\textbf{LLM for Evaluation:} We employ GPT-4o to classify errors in the reasoning traces. The prompts used for evaluation are provided in Appendix~\ref{appendix:prompts}.

\textbf{Evaluation Dataset:} To assess the LLM's classification accuracy for error tagging, we curated a human-annotated dataset. It contains 25 responses for each error type and 100 responses with no errors. Each response with an error tag is annotated with one error type.

\section{Results and Discussion}
\label{sec:results}
Figure~\ref{fig:eval} shows the LLM's evaluation accuracy on human-annotated responses curated as described in experimental setup. The accuracy reported is 57.8\%. with Format Error which has low recall due to conversion to FOL. 

The test dataset for LLM evaluation was partitioned into four patterns based on the correctness of the base model and the RL model: Pattern 00 corresponds to cases where both models produced incorrect final answers; Pattern 01 where the RL Model was correct while the base model was incorrect; Pattern 11 where both models were correct; and Pattern 10, a rare case where the RL model was incorrect while the base model was correct.

\begin{figure}[t]
    \centering

    \begin{subfigure}[b]{0.42\textwidth}
        \centering
        \includegraphics[width=\textwidth]{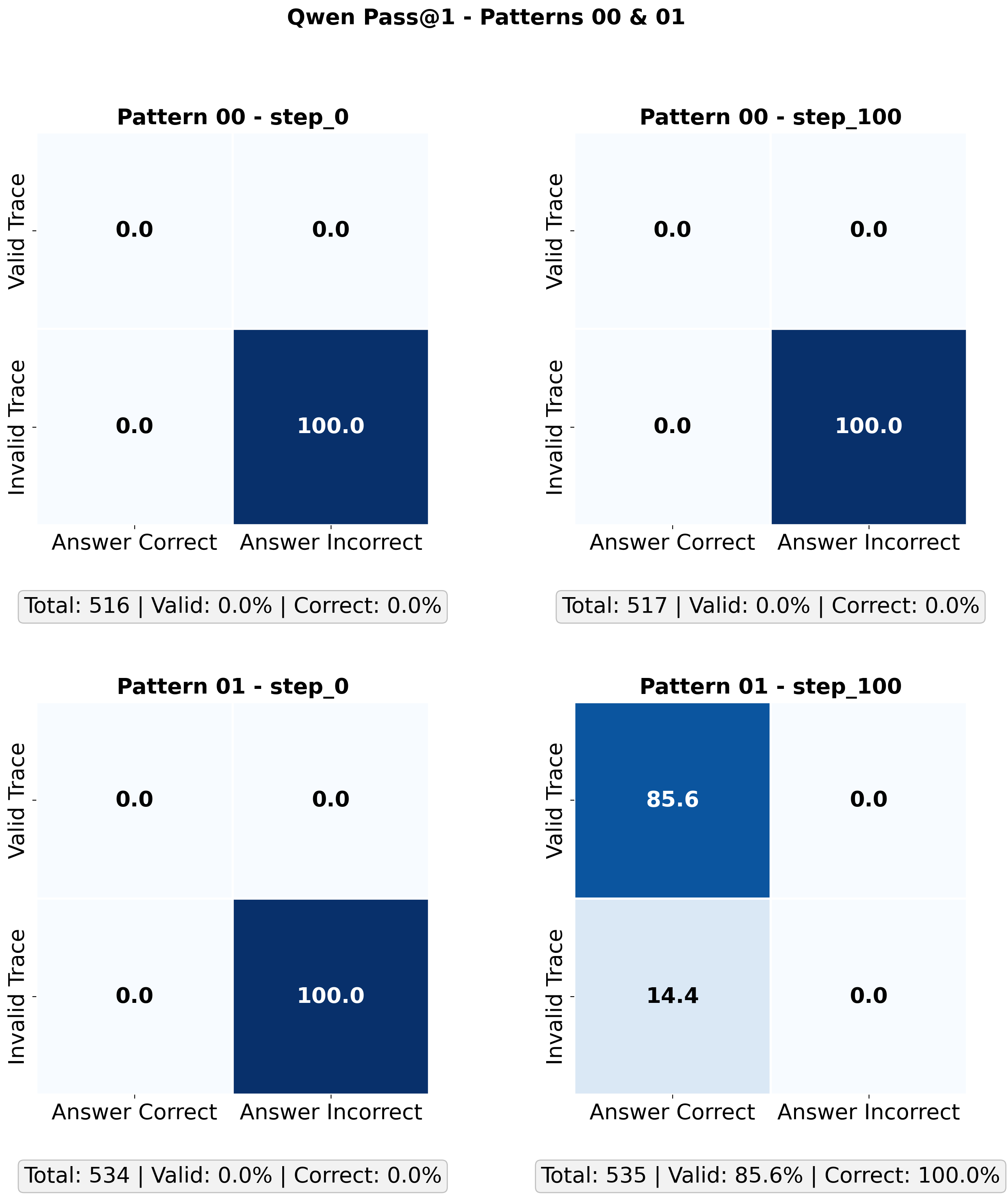}
        \caption{Pattern 00 \& 01, Pass$@$1}
        \label{fig:fig1}
    \end{subfigure}
    \hfill
    \begin{subfigure}[b]{0.42\textwidth}
        \centering
        \includegraphics[width=\textwidth]{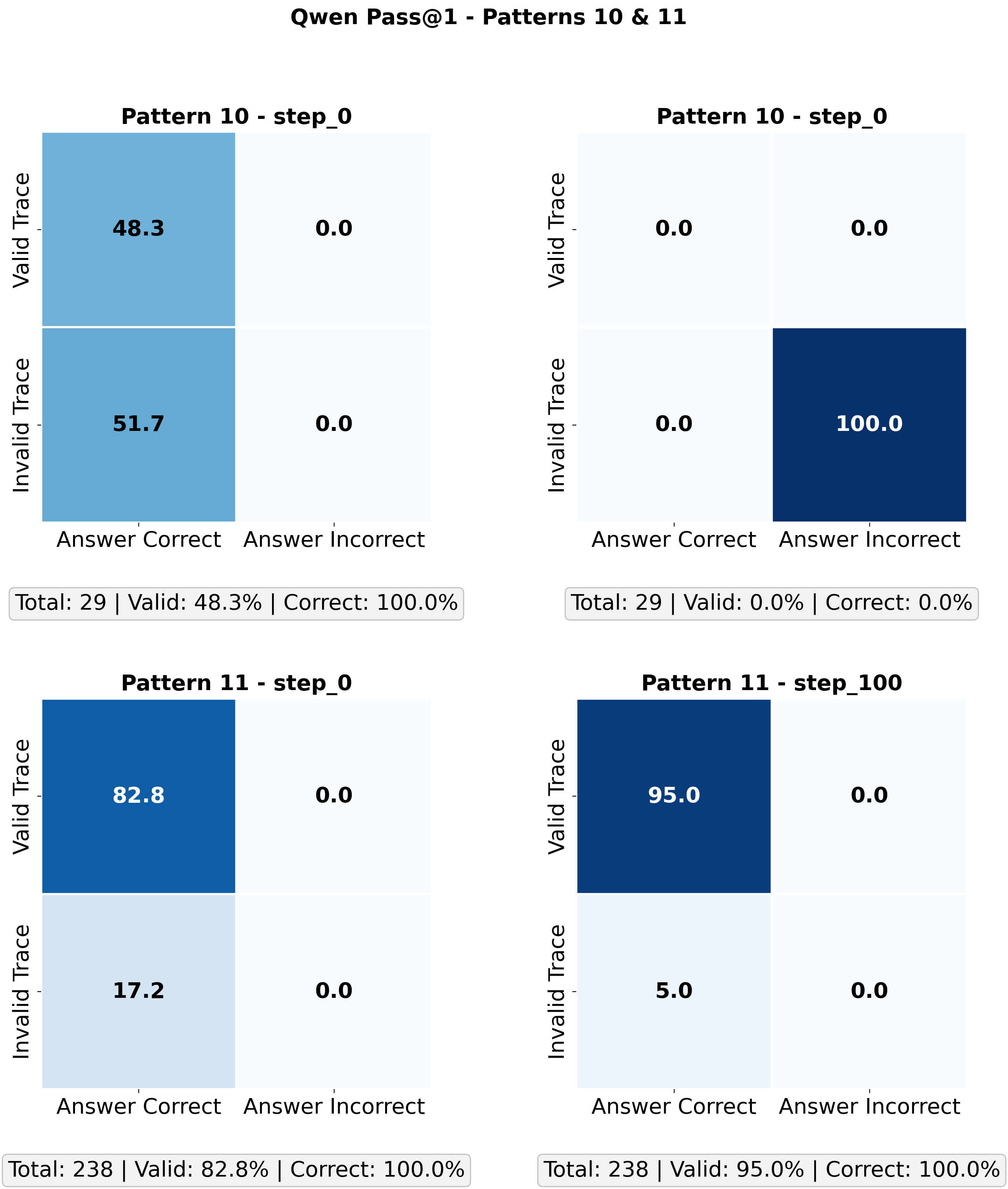}
        \caption{Pattern 10 \& 11, Pass$@$1}
        \label{fig:fig2}
    \end{subfigure}

    \caption{Confusion matrices for patterns 00 \& 01 (\textit{left}) and 10 \& 11 (\textit{right}) at Pass$@$1.}
    \label{fig:pass1}
\end{figure}

Figure~\ref{fig:fig1} presents the confusion matrices for Patterns 00 and 01 for Pass@1. Additional results for Pass@4 and Pass@16 are provided in appendix~\ref{appendix:results}. For Pattern 00, by definition of Pass@K trace coherence, traces become invalid when the final answers are incorrect. In contrast, for Pattern 01, the RL model substantially improves trace coherence, reaching approximately 85\% across all Pass@K values compared to 0\% for the base model. Thus pattern 01 Shows an improvement in trace coherence where the RL model got the final answers correct.

Figure~\ref{fig:fig2} reports results for Patterns 11 and 10 for Pass@1, and results for Pass@4 and Pass@16 could be found in appendix~\ref{appendix:results}. For Pattern 10, coherence is invalid for the RL model as all answers are incorrect across passes while for Pattern 11, the RL model shows a consistent improvement in trace coherence achieving up to 96\% across Pass@K values compared to the base model. Overall, these results indicate that RLVR improved trace coherence in cases where the RL model achieves correct final answers.

\section{Conclusion}
In this work, we investigated the effect of Reinforcement Learning with Verifiable Rewards (RLVR) on the intermediate reasoning steps of Large Language Models (LLMs). While prior studies focused primarily on final answer accuracy, we introduced the concept of \emph{trace coherence} which which is implied by trace validity for cases where formal correctness is infeasible to verify, such as in math word problems. Trace coherence acts as a proxy for trace validity by evaluating the impact of RLVR on reasoning traces using an error taxonomy grounded in First-Order Logic (FOL). By leveraging LLMs-as-a-Judge to classify errors in intermediate steps, we systematically evaluated trace coherence across different Pass@K values on the GSM8K benchmark. Our results demonstrate that RLVR post-training improves trace coherence, particularly in problems where the final answers become correct after RLVR post training. This suggests that RLVR can enhance perceived trace quality through improvements in local coherence. We thus draw a clear distinction that, while RLVR improves trace coherence, it does not amount to trace validity or overall correctness in mathematical reasoning problems. Improvements in trace coherence reflect local consistency but should not be mistaken for improved correctness unless validated through systematic and formal evaluation.

\section*{Acknowledgment}
This research is supported in part by grants from ONR (N00014-25-1-2301 and N00014-23-1-2409), DARPA (HR00112520016), DoD RAI (via CMU subcontract 25-00306-SUB-000), an Amazon Research Award, and a generous gift from Qualcomm.


\bibliography{references}

\begin{thebibliography}{10}

\bibitem{bhambri2025cognitively}
Siddhant Bhambri, Upasana Biswas, and Subbarao Kambhampati.
\newblock Do cognitively interpretable reasoning traces improve llm performance?
\newblock {\em arXiv preprint arXiv:2508.16695}, 2025.

\bibitem{bhambri2025interpretable}
Siddhant Bhambri, Upasana Biswas, and Subbarao Kambhampati.
\newblock Interpretable traces, unexpected outcomes: Investigating the disconnect in trace-based knowledge distillation.
\newblock {\em arXiv preprint arXiv:2505.13792}, 2025.

\bibitem{cobbe2021training}
Karl Cobbe, Vineet Kosaraju, Mohammad Bavarian, Mark Chen, Heewoo Jun, Lukasz Kaiser, Matthias Plappert, Jerry Tworek, Jacob Hilton, Reiichiro Nakano, et~al.
\newblock Training verifiers to solve math word problems.
\newblock {\em arXiv preprint arXiv:2110.14168}, 2021.

\bibitem{cui2025entropy}
Ganqu Cui, Yuchen Zhang, Jiacheng Chen, Lifan Yuan, Zhi Wang, Yuxin Zuo, Haozhan Li, Yuchen Fan, Huayu Chen, Weize Chen, et~al.
\newblock The entropy mechanism of reinforcement learning for reasoning language models.
\newblock {\em arXiv preprint arXiv:2505.22617}, 2025.

\bibitem{gu2024survey}
Jiawei Gu, Xuhui Jiang, Zhichao Shi, Hexiang Tan, Xuehao Zhai, Chengjin Xu, Wei Li, Yinghan Shen, Shengjie Ma, Honghao Liu, et~al.
\newblock A survey on llm-as-a-judge.
\newblock {\em arXiv preprint arXiv:2411.15594}, 2024.

\bibitem{guo2025deepseek}
Daya Guo, Dejian Yang, Haowei Zhang, Junxiao Song, Ruoyu Zhang, Runxin Xu, Qihao Zhu, Shirong Ma, Peiyi Wang, Xiao Bi, et~al.
\newblock Deepseek-r1: Incentivizing reasoning capability in llms via reinforcement learning.
\newblock {\em arXiv preprint arXiv:2501.12948}, 2025.

\bibitem{lewkowycz2022solving}
Aitor Lewkowycz, Anders Andreassen, David Dohan, Ethan Dyer, Henryk Michalewski, Vinay Ramasesh, Ambrose Slone, Cem Anil, Imanol Schlag, Theo Gutman-Solo, et~al.
\newblock Solving quantitative reasoning problems with language models.
\newblock {\em Advances in neural information processing systems}, 35:3843--3857, 2022.

\bibitem{li2024evaluating}
Xiaoyuan Li, Wenjie Wang, Moxin Li, Junrong Guo, Yang Zhang, and Fuli Feng.
\newblock Evaluating mathematical reasoning of large language models: A focus on error identification and correction.
\newblock {\em arXiv preprint arXiv:2406.00755}, 2024.

\bibitem{liu2025prorl}
Mingjie Liu, Shizhe Diao, Ximing Lu, Jian Hu, Xin Dong, Yejin Choi, Jan Kautz, and Yi~Dong.
\newblock Prorl: Prolonged reinforcement learning expands reasoning boundaries in large language models.
\newblock {\em arXiv preprint arXiv:2505.24864}, 2025.

\bibitem{liu2025understanding}
Zichen Liu, Changyu Chen, Wenjun Li, Penghui Qi, Tianyu Pang, Chao Du, Wee~Sun Lee, and Min Lin.
\newblock Understanding r1-zero-like training: A critical perspective.
\newblock {\em arXiv preprint arXiv:2503.20783}, 2025.

\bibitem{marjanovic2025deepseek}
Sara~Vera Marjanovi{\'c}, Arkil Patel, Vaibhav Adlakha, Milad Aghajohari, Parishad BehnamGhader, Mehar Bhatia, Aditi Khandelwal, Austin Kraft, Benno Krojer, Xing~Han L{\`u}, et~al.
\newblock Deepseek-r1 thoughtology: Let's think about llm reasoning.
\newblock {\em arXiv preprint arXiv:2504.07128}, 2025.

\bibitem{samineni2025rl}
Soumya~Rani Samineni, Durgesh Kalwar, Karthik Valmeekam, Kaya Stechly, and Subbarao Kambhampati.
\newblock Rl in name only? analyzing the structural assumptions in rl post-training for llms.
\newblock {\em arXiv preprint arXiv:2505.13697}, 2025.

\bibitem{sheng2025hybridflow}
Guangming Sheng, Chi Zhang, Zilingfeng Ye, Xibin Wu, Wang Zhang, Ru~Zhang, Yanghua Peng, Haibin Lin, and Chuan Wu.
\newblock Hybridflow: A flexible and efficient rlhf framework.
\newblock In {\em Proceedings of the Twentieth European Conference on Computer Systems}, pages 1279--1297, 2025.

\bibitem{Stechly2025BeyondST}
Kaya Stechly, Karthik Valmeekam, Atharva Gundawar, Vardhan Palod, and Subbarao Kambhampati.
\newblock Beyond semantics: The unreasonable effectiveness of reasonless intermediate tokens.
\newblock {\em ArXiv}, abs/2505.13775, 2025.

\bibitem{wei2022chain}
Jason Wei, Xuezhi Wang, Dale Schuurmans, Maarten Bosma, Fei Xia, Ed~Chi, Quoc~V Le, Denny Zhou, et~al.
\newblock Chain-of-thought prompting elicits reasoning in large language models.
\newblock {\em Advances in neural information processing systems}, 35:24824--24837, 2022.

\bibitem{yu2025dapo}
Qiying Yu, Zheng Zhang, Ruofei Zhu, Yufeng Yuan, Xiaochen Zuo, Yu~Yue, Tiantian Fan, Gaohong Liu, Lingjun Liu, Xin Liu, et~al.
\newblock Dapo: An open-source llm reinforcement learning system at scale.
\newblock {\em arXiv preprint arXiv:2503.14476}, 2025.

\bibitem{yue2025does}
Yang Yue, Zhiqi Chen, Rui Lu, Andrew Zhao, Zhaokai Wang, Shiji Song, and Gao Huang.
\newblock Does reinforcement learning really incentivize reasoning capacity in llms beyond the base model?
\newblock {\em arXiv preprint arXiv:2504.13837}, 2025.

\bibitem{zhang2025grpo}
Jixiao Zhang and Chunsheng Zuo.
\newblock Grpo-lead: A difficulty-aware reinforcement learning approach for concise mathematical reasoning in language models.
\newblock {\em arXiv preprint arXiv:2504.09696}, 2025.

\end{thebibliography}
\bibliographystyle{plain}

\appendix

\section{Prompt for error categorization for GSM8K Dataset}
\label{appendix:prompts}
\begin{tcolorbox}[colback=black!5!white, colframe=black!75!black,
                  title=ERROR TAGGING PROMPT, breakable]

1. Read the problem carefully and understand what is being asked.\\
2. Verify the model response by converting it into First-Order Logic (FOL) and identifying any errors based on the taxonomy below.\\
3. You MUST return a valid JSON object with exactly these three keys: 
   \texttt{"First-Order Logic"}, \texttt{"error tags"}, \texttt{"rationale"}.

\textbf{ERROR TAXONOMY (use these exact labels):}

\begin{enumerate}
  \item \textbf{False Premise} -- The model makes incorrect assumptions or misinterprets the problem:
    \begin{itemize}
      \item Assumes data not given in the problem
      \item Misinterprets key terms or relationships
      \item Ignores important given information
      \item Uses wrong units or constants (e.g., week $\neq$ 7 days)
    \end{itemize}

  \item \textbf{False Rule} -- The model applies incorrect mathematical logic:
    \begin{itemize}
      \item Uses wrong formulas or operations
      \item Applies operations to incompatible types
      \item Makes logical inference errors
      \item Misunderstands mathematical relationships
      \item Omits intermediate calculations needed for the answer
      \item Only flag if the step is truly necessary for correctness
    \end{itemize}

  \item \textbf{Calculator Error} -- The model makes wrong addition, subtraction, multiplication, or division:
    \begin{itemize}
      \item ONLY flag if calculator arithmetic is wrong (e.g., $5 \times 6 = 10$ instead of 30)
      \item Do NOT flag if calculations are correct but final answer differs from expected
      \item Do NOT flag if correct math is applied to wrong numbers
    \end{itemize}

    \item \textbf{Format Error} 
    \begin{itemize}
      \item ONLY flag if the final answer is not within boxed.
    \end{itemize}

  \item \textbf{None} -- No errors found (use empty list []):
    \begin{itemize}
      \item All calculations are correct
      \item All necessary steps shown
      \item Correct interpretation of problem
    \end{itemize}
\end{enumerate}

\textbf{Examples:}
1. Problem: "In a survey, 40\% of students said they like basketball, 30\% said they like chess, and 20\% said they like both. If 200 students were interviewed, how many students like either basketball or chess?" \\
Model Response: "Number of students who like basketball = 200 × 0.4 = 80 \\
Number of students who like chess = 200 × 0.3 = 60 \\
Total students who like basketball or chess = 80 + 60 = 140 \\boxed{140}" \\
error\_tags: ["False Rule"] \\
rationale: The model incorrectly assumes the sets of basketball and chess fans are disjoint. Since 20\% of students like both sports, the correct total should subtract the overlap: \\
80 + 60 − 40 = 100. Adding the numbers directly violates the rules of set union, leading to an overcount.

\vspace{1em}
\noindent
2. Problem:"Today, at the school science lesson, Jake learned that there are 8 more solar systems for every planet in the galaxy. If there are 20 planets in the galaxy, how many solar systems and planets are there altogether?" \\
Model Response: "The galaxy has 20 planets. Each planet has 8 solar systems. Total solar systems = 20 × 9 = 180. Total planets + solar systems = 180 + 20 = 200. \\boxed{200}" \\
error\_tags: ["False Premise"] \\
rationale: "The model misinterprets the problem by considering 9 more solar systems for every planet' as '9 solar systems per planet instead of 8, given in the problem"

\vspace{1em}
\noindent
3. Problem: "If Alice purchased 3600 acorns for nine times the price Bob paid, and Bob paid \$6000 for his acorns, how much did Alice pay for each acorn?" \\
Model Response:"Total Alice paid = 9 × Total Bob paid = 9 × 6000 = 54000 \\
Price per acorn for Alice = 15 \\boxed{15}" \\
error\_tags: ["False Rule"] \\
rationale: "The model correctly computes the total amount Alice paid but skips the division step that shows how the per-acorn price is calculated: Price per acorn for Alice = 54000 / 3600 = 15. This missing step is necessary to make the logical reasoning complete."

\vspace{1em}
\noindent
4. Problem: "Erica sees 9 butterflies in the garden. She sees one-third of them fly away. How many butterflies are left in the garden?" \\
Model Response: "One-third of 9 = 4. Remaining butterflies = 9 - 4 = 5. \\boxed{5}" \\
error\_tags: ["Calculator Error"] \\
rationale: "The calculation 'one-third of 9 = 4' is mathematically wrong. It should be 9 ÷ 3 = 3. This is an actual calculator error."

\vspace{1em}
\noindent
5. Problem: "5 geckos eat 6 insects each. 3 lizards eat twice as much as the geckos. How many total insects were eaten?" \\
Model Response: "Geckos eat: 5 × 6 = 30 insects. Lizards eat: 3 × 12 = 36 insects. Total: 30 + 36 = 66 insects. \\boxed{66}" \\
error\_tags: [] \\
rationale: "All arithmetic calculations are mathematically correct: 5 × 6 = 30, 3 × 12 = 36, and 30 + 36 = 66. No arithmetic errors present. GIven data is used correctly. No False premise, rule applied properly and all steps are used correctly."

\vspace{1em}
\noindent
7. Problem: "There are 4 tables with 4 legs each, 1 sofa, 2 chairs with 4 legs each, 3 tables with 3 legs each, 1 table with 1 leg, and 1 rocking chair with 2 legs. How many legs total?" \\
Model Response: "4 tables × 4 legs = 16, 1 sofa × 4 legs = 4, 2 chairs × 4 legs = 8, 3 tables × 3 legs = 9, 1 table × 1 leg = 1, 1 rocking chair × 2 legs = 2. Total: 16 + 4 + 8 + 9 + 1 + 2 = 40 legs. \\boxed{40}" \\
error\_tags: [] \\
rationale: "The model makes reasonable assumptions (sofa has 4 legs) based on common knowledge. All calculations are correct: 4×4=16, 1×4=4, 2×4=8, 3×3=9, 1×1=1, 1×2=2, total=40. No errors present."

\vspace{1em}
\noindent
8. Problem: "If Alice purchased 3600 acorns for nine times the price Bob paid, and Bob paid \$6000 for his acorns, how much did Alice pay for each acorn?" \\
Model Response: "Bob's price per acorn = \$6000 ÷ b. Alice's total payment = 9 × \$6000 = \$54,000. Alice's price per acorn = \$54,000 ÷ 3600 = \$15. \\boxed{15}" \\
error\_tags: [] \\
rationale: "The model correctly calculates Alice's price per acorn. While it mentions Bob's price per acorn calculation (which is not needed), it still arrives at the correct final answer: \$54,000 ÷ 3600 = \$15. No errors present."
\textbf{OUTPUT FORMAT (JSON):}
\begin{verbatim}
{
    "First-Order Logic": "Signature: ...\nFormalization: ...\nDerivation: 
    ...\nCheck: ...",
    "error_tags": ["error_type"],
    "rationale": "Brief explanation for each error tag applied"
}
\end{verbatim}

\textbf{Instructions:}
1) Convert the given model response into First-Order Logic with sections: Signature (variables/constants), Formalization (First order Logic statements), Derivation (logical steps), Check (verification).\\

2) Compare the model's FOL against the problem requirements. Identify where the premises contradict the problem, rules are invalid, steps are missing, or arithmetic is incorrect.\\

3) For \textbf{False Premise}: Only apply if the model makes unreasonable assumptions that contradict the problem or common sense
\begin{itemize}
   \item Do NOT apply False Premise for reasonable inferences based on common knowledge (e.g., assuming standard furniture has typical leg counts)
   \item Do NOT apply False Premise when the model correctly identifies that certain information is not needed for the solution
   \item Do NOT apply False Premise when the model correctly sets up mathematical relationships and solves them properly
\end{itemize}

4) For \textbf{False Rule}: Only apply if the model uses an incorrect mathematical rule or operation (e.g., adding percentages incorrectly, using wrong formulas)
\begin{itemize}
   \item Do NOT apply False Rule if the model correctly follows the problem's mathematical requirements
\end{itemize}

5) For \textbf{Calculator Error}:
\begin{itemize}
   \item Only flag if addition, subtraction, multiplication and division are incorrect (e.g., $5 \times 6 = 10$, is incorrect multiplication)
   \item Do NOT apply calculator error tag if the addition, subtraction, multiplication and division are correct
\end{itemize}

6) For \textbf{``None''} error tag: Only return empty \texttt{error\_tags []} if the model response is completely correct with no errors of any type.
\begin{itemize}
   \item A response with correct calculations, complete steps, and accurate premises should have \texttt{error\_tags: []}
   \item Do NOT return ``None'' as an error tag -- use an empty list \texttt{[]} instead
   \item Only flag errors when they actually exist
\end{itemize}

7) Output ALL applicable \texttt{error\_tags} from the allowed set, only if there are errors. Return empty list \texttt{[]} if there are no errors.\\

8) Provide a brief rationale for each error tag applied.\\[1em]

\textbf{CRITICAL RULES:}
\begin{itemize}
   \item Use exact error labels: \texttt{"False Premise"}, \texttt{"False Rule"}, \texttt{"Calculator Error"}.
   \item For no errors, use empty list: \texttt{[]}
   \item Only flag errors that actually exist
   \item Be conservative -- when in doubt, don't flag an error
   \item Focus on the most obvious/clear errors first
   \item \textbf{CALCULATOR ERROR RULE}: Only flag if the actual math is wrong (e.g., $5\times 6 = 10$). Do NOT flag if calculations are correct but the answer differs from expected
\end{itemize}

\textbf{IMPORTANT:} Return ONLY a valid JSON object. Do not include any other text before or after the JSON.

\end{tcolorbox}
\section{Error Categories description with Examples}
\label{appendix:error}
\begin{enumerate}
  \item \textbf{False Premise:}
    \begin{itemize}
      \item \textbf{Conceptual Misunderstanding:} Misinterpreting the overall scenario or problem structure.
            Example: Reading ``4 vacations per year'' as ``4 vacations total''.
      \item \textbf{Semantic Error:} Misusing terms or values that distort the intended meaning.
            Example: Treating a monthly salary of \$600 as annual.
      \item \textbf{False Assumption:} Introducing facts not supported by the problem.
            Example: Assuming ``the pineapple drink is spilled'' when only ``a drink'' is mentioned.
      \item \textbf{Units Misinterpretation:} Confusing or conflating measurement units or quantities.
            Example: Interpreting a tank's total volume as water poured.
   
    \end{itemize}

  \item \textbf{False Rule:}
    \begin{itemize}
      \item \textbf{Type / Operand Mismatch:} Operation applied to incompatible types, units, or domains.
            Example: Adding 10\% of monthly salary to compute an annual raise.
      \item \textbf{Inference Violation:} Conclusion does not logically follow from valid premises.
            Example: From ``some cats are black,'' infer ``all black things are cats.''
      \item \textbf{Operation Misapplication:} The operation or rule itself is inappropriate for the problem context.
            Example: Using compound interest for a one-time insurance fee.
      \item \textbf{Quantifier Misuse:} Misplacing or misinterpreting logical quantifiers $\forall$ and $\exists$,
            causing overgeneralization or unwarranted restriction. Example: Interpreting ``each'' as ``all at once.''
    \item \textbf{Missing necessary steps:} Key reasoning or computation steps are skipped, breaking the logical chain to reach a valid conclusion.
    \end{itemize}

  \item \textbf{Calculator Error:}
    \begin{itemize}
      \item Simple numeric or arithmetic mistakes \textbf{only if the numeric calculation is mathematically incorrect}.
      \item Examples: \(5 \times 6 = 10\) instead of \(30\); miscomputing \(7.5 + 2.5 = 11\); \(9 \div 3 = 4\) instead of \(3\).
    \end{itemize}
\item \textbf{Format Error:}
    The final answer to be formatted according to the instruction in the \\boxed
    
\end{enumerate}

\section{Training Hyper-parameters}
\label{appendix:hyper_params}
The training batch size is set to 64, with a mini-batch size of 8. We sample 5 responses per question prompt. During training, the response rollouts for each question are generated with a temperature of 0.6., The maximum prompt length is set to 512 for GSM8K, while the maximum response length is fixed at 1024. The learning rate is set to 1e-6. And, we set the KL divergence coefficient to $\beta = 1e-3$. All experiments on the GSM8K dataset are conducted using a single A100 80GB GPU. Both the Qwen and Llama family models are trained for 145 global time steps, corresponding to 5 epochs. Model evaluation is performed at three different time steps (0, 10, and 100) on the GSM8K test dataset, using Pass@K ($k=1,4,16$). For response sampling during evaluation, we set the temperature to 1.0 and top-p to 0.95.

The source code will be provided on acceptance.

\section{Additional Results}
\label{appendix:results}
\begin{figure}[htbp]
    \centering

    \begin{subfigure}[b]{0.3\textwidth}
        \centering
        \includegraphics[width=\textwidth]{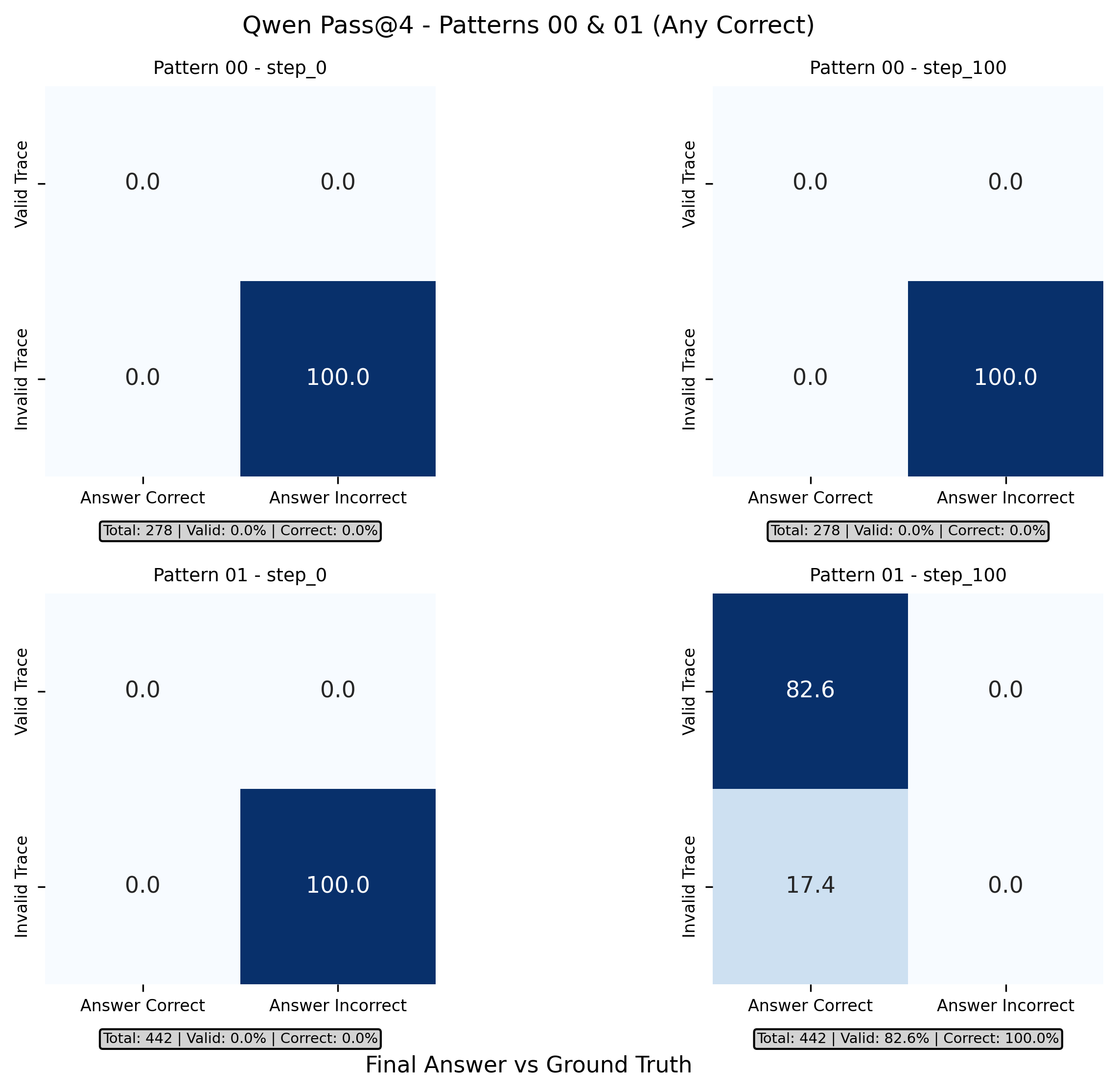}
        \caption{Pattern 00 \& 01, pass$@$4}
        \label{fig:fig3}
    \end{subfigure}
    \hspace{0.02\textwidth}
    \begin{subfigure}[b]{0.3\textwidth}
        \centering
        \includegraphics[width=\textwidth]{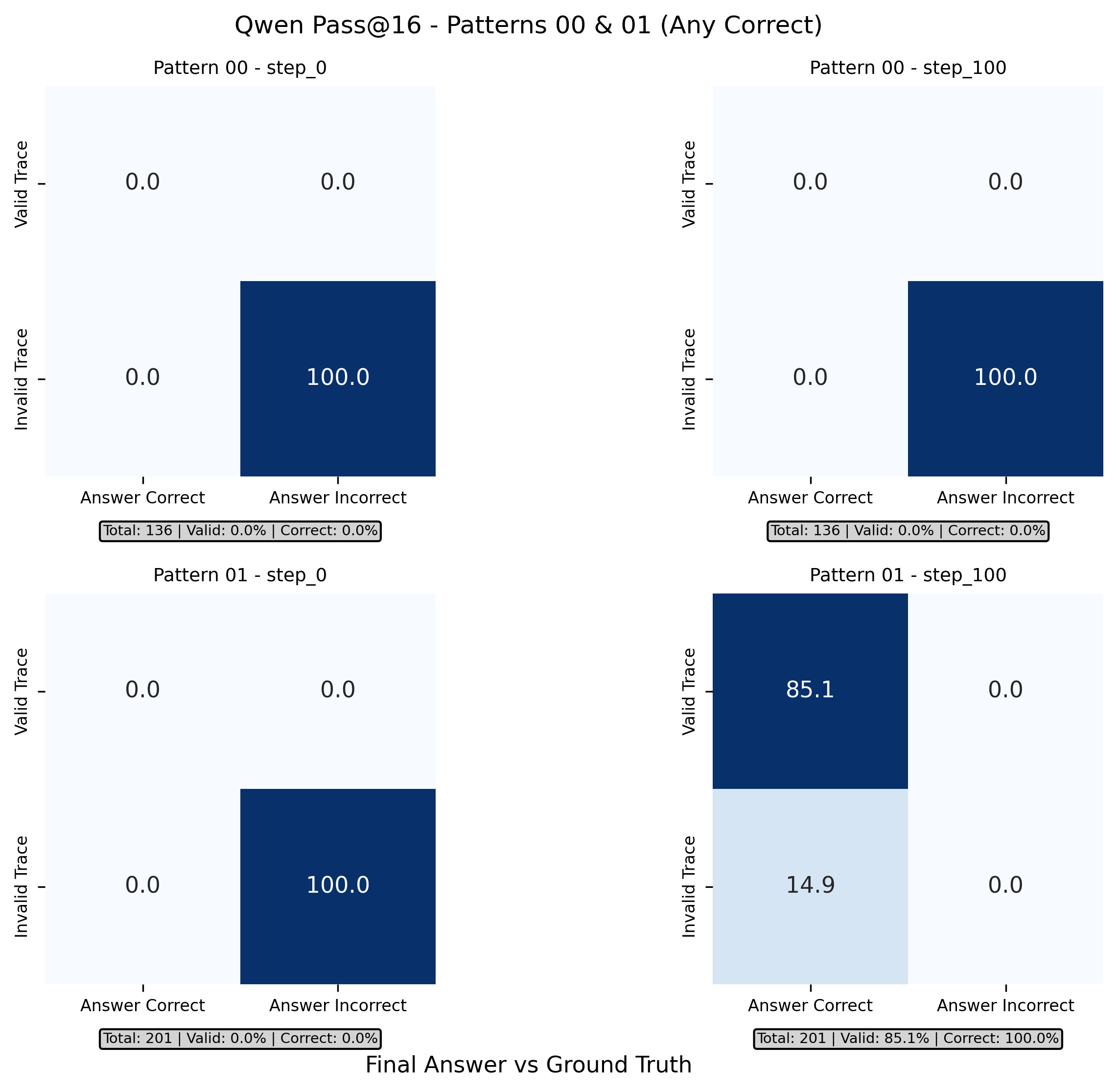}
        \caption{Pattern 00 \& 01, pass$@$16}
        \label{fig:fig4}
    \end{subfigure}

    \vspace{0.5em} 

    \begin{subfigure}[b]{0.3\textwidth}
        \centering
        \includegraphics[width=\textwidth]{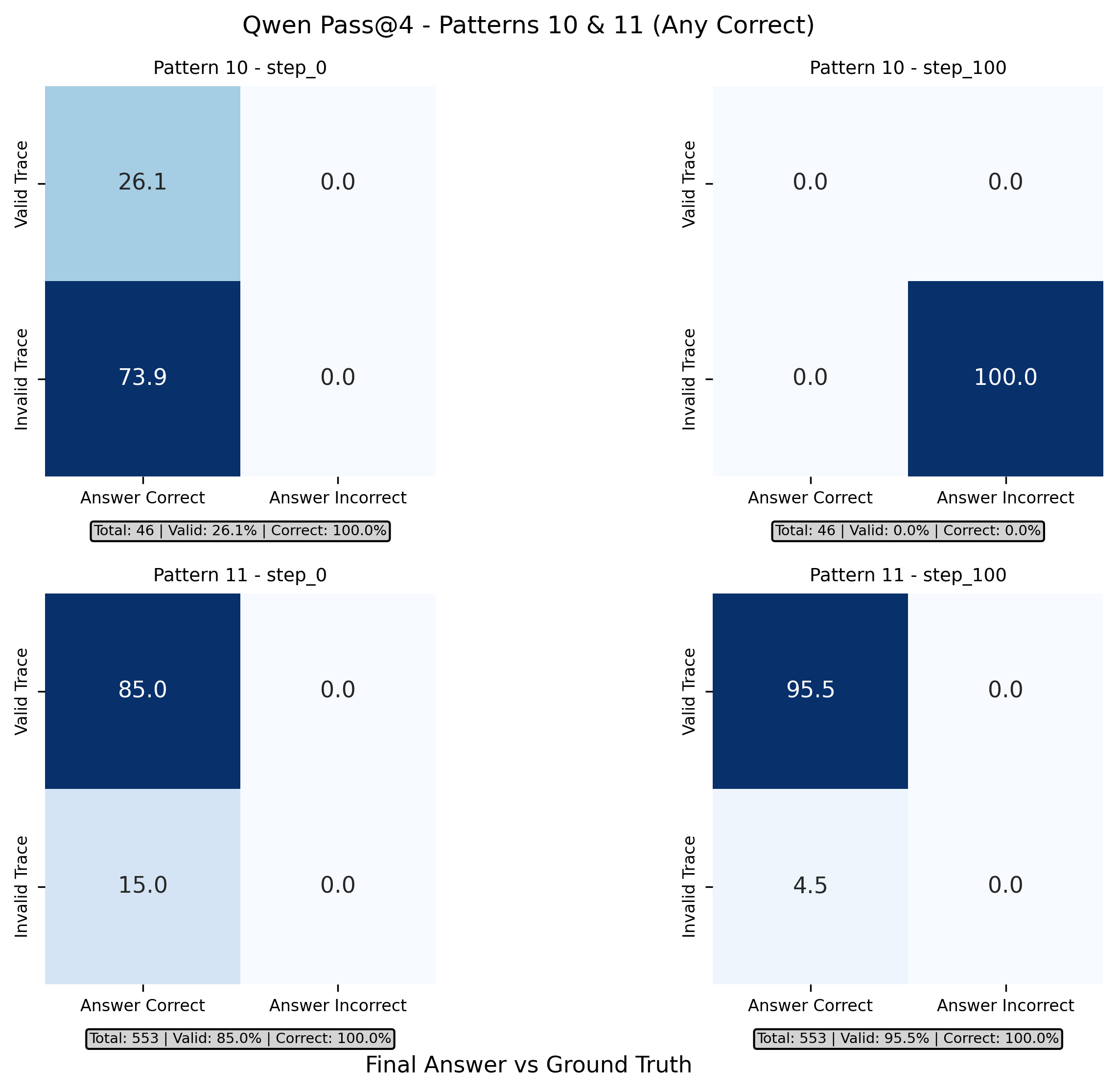}
        \caption{Pattern 10 \& 11, pass$@$4}
        \label{fig:fig5}
    \end{subfigure}
    \hspace{0.02\textwidth}
    \begin{subfigure}[b]{0.3\textwidth}
        \centering
        \includegraphics[width=\textwidth]{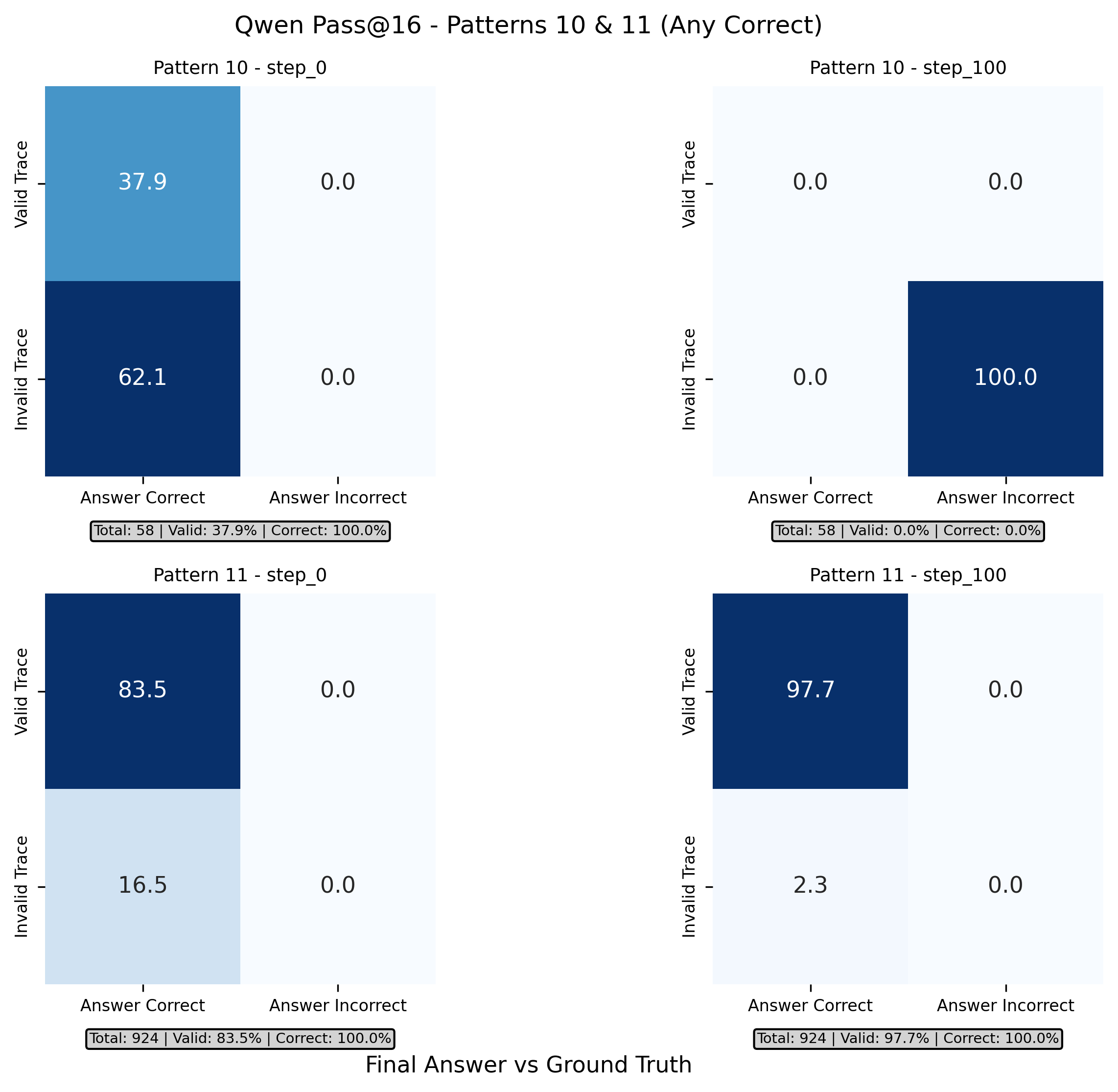}
        \caption{Pattern 10 \& 11, pass$@$16}
        \label{fig:fig6}
    \end{subfigure}

    \caption{Confusion matrices for patterns 00, 01, 10 and 11 at pass$@4$ and pass$@16$}
    \label{fig:combined_confusion_matrices}
\end{figure}

\end{document}